# An Enhancement of Haar Cascade Algorithm Applied to Face Recognition for Gate Pass Security

**Clarence A. Antipona[1], Romeo R. Magsino III[2], Raymund M. Dioses[3] , Khatalyn E. Mata[4]**

[1] Computer Science Department, Pamantasan ng Lungsod ng Maynila, Manila, Philippines.

[2] Computer Science Department, Pamantasan ng Lungsod ng Maynila, Manila, Philippines.

[3] Computer Science Department, Pamantasan ng Lungsod ng Maynila, Manila, Philippines.

[4] Computer Science Department, Pamantasan ng Lungsod ng Maynila, Manila, Philippines.

**Email:** [1]antiponaclarence@gmail.com, [2]rirmagsino2021@plm.edu.ph, [3]rmdioses@plm.edu.ph, [4]kemata@plm.edu.ph

**Orchid Id number:** [1]0009-0009-3834-219X

**Corresponding Author*:** Clarence A. Antipona

**ABSTRACT:**

*This study is focused on enhancing the Haar Cascade Algorithm to decrease the false positive and false negative rate in face matching and face detection to increase the accuracy rate even under challenging conditions. The face recognition library was implemented with Haar Cascade Algorithm in which the 128-dimensional vectors representing the unique features of a face are encoded. A subprocess was applied where the grayscale image from Haar Cascade was converted to RGB to improve the face encoding. Logical process and face filtering are also used to decrease non-face detection. The Enhanced Haar Cascade Algorithm produced a 98.39% accuracy rate (21.39% increase), 63.59% precision rate, 98.30% recall rate, and 72.23% in F1 Score. In comparison, the Haar Cascade Algorithm achieved a 46.70% to 77.00% accuracy rate, 44.15% precision rate, 98.61% recall rate, and 47.01% in F1 Score. Both algorithms used the Confusion Matrix Test with 301,950 comparisons using the same dataset of 550 images. The 98.39% accuracy rate shows a significant decrease in false positive and false negative rates in facial recognition. Face matching and face detection are more accurate in images with complex backgrounds, lighting variations, and occlusions, or even those with similar attributes.*

1) **Introduction:**

Facial recognition is increasingly common in various applications ranging from security systems to social media platforms. The effectiveness of such systems hinges upon the robustness of the underlying algorithms employed for facial recognition. Haar Cascade algorithms have been widely utilized in this domain due to their computational efficiency and satisfactory accuracy. Haar Cascade algorithm is a type of machine learning wherein a classifier is used from a great deal of positive and negative photos. Haar feature-based cascade classifiers are utilized for object detection purposes. This classifier employs a machine learning technique where cascade operation is applied to images to detect objects in subsequent images [7].

However, challenges persist in achieving reliable facial recognition under various conditions such as changes in lighting, facial expressions, and occlusions. Typical facial recognition doesn't go into the depth of the image, it just checks for the relevant similarities. This can sometimes become a vulnerability to bypass this kind of system [4]. Before the introduction of the Haar Cascade algorithm in 2001, several object recognition applications were developed. Additional principal component analysis (PCA) was utilized to reduce the complexity of face images, decrease data size, and eliminate noise. PCA was also integrated with neural networks for face recognition and sex determination. However, these earlier algorithms exhibited certain drawbacks, such as a low classification percentage (ranging from 31.48% to 94.5%) and high mean square error (ranging from 0.02 to 0.12). A refinement of the Haar cascade algorithm was proposed, which combined three different classifiers: color HSV, histogram matching, and eyes/mouth detection [2]. This enhanced algorithm was subsequently utilized by Arreola in 2018, who applied it to a quad-rotor Unmanned Aerial Vehicle (UAV) for face recognition purposes [2]. In addition to the Haar cascade algorithm, other approaches can be applied to real-time tracking tasks, including local binary patterns (LBPs) or histogram of oriented gradients (HOG). A comparative study involving three algorithms—Haar cascade, LBP, and HOG—was conducted for object detection using UAVs. The results indicated that the Haar-like cascade outperformed LBP in accuracy rate and was faster than HOG [5].

In this study, we will implement a combination of a Haar cascade algorithm, face recognition model from OpenCV, and logical processing for filtering. The Enhancement of Haar Cascade Algorithm Applied to Face Recognition for Gate Pass Security will significantly improve security by providing accurate and efficient identification of individuals, thereby preventing unauthorized access and ensuring the safety of individuals.

## 2) Literature Review:

The Haar Cascade Algorithm, developed by Paul Viola and Michael Jones, are well-known for their significant contribution to object detection. Their innovative approach has greatly influenced subsequent research in the field, including the work described in the paper "Rapid object detection using a boosted cascade of simple features". By introducing efficient methods for feature computation, such as the "integral image" representation, and utilizing learning algorithms like AdaBoost [8]. According to the study "Use of Haar Cascade Classifier for Face Tracking System in Real Time Video." they introduced a comprehensive face detection and tracking system designed for real-time video inputs, emphasizing its application in security contexts [3]. Incorporating the Haar-Cascade method and OpenCV libraries, the system's initial phase encompasses face recognition and detection. Subsequently, face tracking is performed using a Face clustering algorithm. The system primarily targets security purposes, operating on video recordings from public areas to identify and track individuals or suspicious activities.

The paper "A Novel Real-Time Face Detection System Using Modified Affine Transformation and Haar Cascades" addresses the challenges posed by tilted, occlusion, and varying illumination in computer vision. Their approach employed a Haar-cascade classifier enhanced with Modified Census Transform features, traditionally inadequate for detecting faces under such conditions [6]. In the study "Analyzing of Different Features Using Haar Cascade Classifier." it discusses the utilization of face recognition for enhancing security measures focusing on the application of a Haar cascade classifier. Their research aims to investigate the effectiveness of using the Haar cascade classifier for analyzing different features, particularly in the context of identifying and matching faces for access control purposes. Through experimental testing, the study demonstrates the efficacy of the proposed method in accurately processing the features of objects, particularly in the context of facial recognition for room security applications [9].

The study "Face Recognition based Attendance System using Haar Cascade and Local Binary Pattern Histogram Algorithm" They developed a robust attendance system based on face recognition to track student attendance accurately. In their study, they addressed some of the false positives by incorporating a strong threshold based on Euclidean distance values during the detection of unknown individuals. Comparative analysis revealed the superiority of the Haar cascade algorithm with Local Binary Pattern Histogram (LBPH) algorithm over other Euclidean distance-based algorithms like Eigenfaces and Fisherfaces [1]. The Haar cascade was chosen for face detection due to its robustness, while the LBPH algorithm was employed for face recognition, renowned for its robustness against monotonic grayscale transformations. The system achieved a student face recognition rate of 77% with a false positive rate of 28%. Even in scenarios where students wore glasses or had facial hair, the system exhibited remarkable performance [1]. Additionally, the recognition rate for unknown individuals reached nearly 60%, with false positive rates of 14% and 30% with and without applying a threshold, respectively. This highlights the significance of Haar cascade in developing efficient face recognition-based attendance systems.

### 3) **Methodology:**

To address the inaccuracy of the Haar Cascade Algorithm particularly in scenarios with variations in lighting, facial expressions, and occlusions that result in False Positive or False Negative. The researchers applied open-source face_recognition library. The face_recognition library is a popular Python package used for recognizing, and analyzing faces in images it computes face encodings, which are 128-dimensional vectors representing the unique features of a face. Similarity is based on the Euclidean distance between these feature vectors.

1. Calculate the Euclidean distance of face encoding. Each image has their own unique 128 face encoding.
2. Given two face encodings:
    a. Face encoding 1: $E_1 = [E_{1,1},\ E_{1,2}, \ldots, E_{1,128}]$
    b. Face encoding 2: $E_2 = [E_{2,1},\ E_{2,2}, \ldots, E_{2,128}]$
3. For two face encodings, $E_1$ and $E_2$, the Euclidean distance $d_{face}$ is:

$$d_{face} = \sqrt{\sum_{i=1}^{128}(E_{1,i} - E_{2,i})^2}$$

- $d_{face}$ = is the distance of the two face encodings.
- $\sum_{i=1}^{128}(E_{1,i} - E_{2,i})^2$ = summation of the 128 face encodings from the images.

4. The similarity percentage can be calculated as:

$$Similarity\ Percentage = \left(1 - \frac{d_{face}}{d_{max}}\right) \times 100 + 25 \quad (1)$$

Where:
  If d_face = 0, it indicates a perfect match, resulting in 100% similarity.
  If d_face = d_max, it indicates a not match, resulting in 0% similarity.
  d_max = 1.0, is the assumed maximum distance.
5. With this calculation, we can assume the similarity percentage of the two images, if the similarity percentage is ≥ 75% then the images is taken as match, if not then it's not match.

To enhance the algorithm that is more reliable in facial matching in faces that share similar attributes (such as wearing glasses). The researchers applied RGB conversion with face encoding. face_encoding will compute the 128-dimensional vectors from RGB images, and not grayscale images. The following are the steps to convert the image to RGB while retaining the function of Haar Cascade Algorithm for grayscale image.

1. Haar Cascade Algorithm will convert the image to grayscale for face detection.
2. The coordinates of the detected face from the grayscale image will be saved.
3. The grayscale image will be converted to RGB.
4. The RGB image will use the saved coordinates where the face is detected.
5. Finally, the coordinates where the face is located will be encoded using the face_encoding where the 128-dimensional vectors are extracted.

To detect real faces the researchers will utilize the Haarcascade pre-trained XML file and enhancement to the detectMultiScale method, specifically adjusting parameters such as scaleFactor, minNeighbors, and minSize using logic process. The result will be filtered by calculating the proximity of detected faces to the center of the image. By determining the image center and calculating the distance of each detected face's center to this point, the process ensures that faces closest to the center are prioritized, thereby reducing the likelihood of false positives and enhancing the accuracy of real face detection.

**Enhanced Haar Cascade Algorithm Pseudocode**

1$^{st}$ Step:
- Load images using OpenCV's imread function.

2$^{nd}$ Step:
- Convert image to grayscale.
- Detect faces using Haar cascade classifier in grayscale images. (scaleFactor, minNeighbors, and minSize are used for face detection.)
- Initialize scaleFactor to 1.1 and minNeighbors to 10, then loop until scaleFactor = 1.01 and minNeighbors = 1.
- Select the largest detected face that is close to the center of the image.
- Put boxes in the detected faces.

3$^{rd}$ Step:
- Convert the detected faces from grayscale to RGB.
- Extract the face encodings of each face using face_recognition and face_encoding.

4$^{th}$ Step:
- Compute the Euclidean distance between the two extracted encodings.
- Convert the distance to a similarity percentage.

5$^{th}$ Step:
- Show images and their similarity percentage.

The Enhanced Haar Cascade Algorithm combines the strengths of the Haar Cascade Algorithm and the face_recognition library to perform a comprehensive analysis of facial features in two distinct images. This process is initiated by the ingestion of the images, followed by a conversion to the RGB format necessary for the operations of the face_recognition library. The core of this process uses the Haar Cascade classifier, a machine learning tool for detecting faces. It looks for facial features like eyes, nose, and mouth by recognizing their spatial arrangement. The classifier works on grayscale images and uses the detectMultiScale function to find possible faces. Instead of manually adjusting the parameters (scaleFactor, minNeighbors, minSize) for each image, we automated the process. The logic iterates through different parameter values and selects the best setting based on filtering. Since faces are usually larger than non-face objects, we choose the largest detected face that is closest to the center of the image as the real face. Once the facial regions are successfully identified, the methodology proceeds with converting these grayscale regions back to RGB format, the RGB image will be processed by face_recognition library. This library utilizes advanced deep learning models to generate numerical 128- dimension encodings that encapsulate the unique features of the detected faces. These encodings from face_encoding, derived from both images, are then subjected to a comparative analysis to determine the degree of similarity between the faces. The comparative analysis hinges on calculating a similarity percentage based on the distance between the facial encodings. A critical element in this process is the establishment of a similarity threshold. This threshold, which can be adjusted to meet specific requirements, plays a pivotal role in classifying pairs of faces as either matching or non-matching. The flexibility of this threshold allows for the modulation of the sensitivity of the facial matching process, making it adaptable to different application contexts. In essence, this process synergistically combines the Haar Cascade's facial detection capabilities with the deep learning-driven encoding and comparison functions of the face_recognition library.

**System Architecture**

Figure 1 illustrates the system architecture of the facial recognition system, which utilizes the enhanced Haar Cascade algorithm. The process begins at the registration stage, where each person's image and personal information are collected and submitted. Each image is processed by the enhanced Haar Cascade algorithm to generate encodings, which are then stored in the database along with the individual's information. In real-time recognition, the system captures live video from a webcam. Each video frame is processed by the Haar Cascade algorithm, generating an encoding that is compared to the stored encodings in the database. If a match is found, the system displays the person's "recognized id" alongside their stored information. If no match is found, the individual is classified as "unknown," and the system sends an alert notification to the appropriate authority. This process ensures real-time identification and alerts for unregistered individuals.

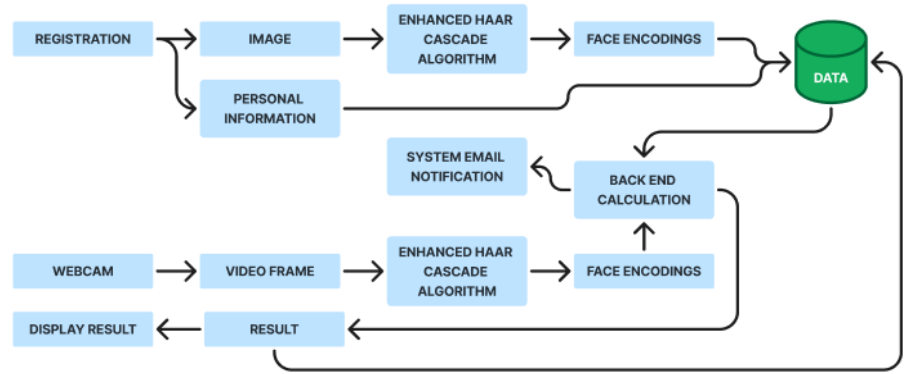

**Figure 1** System Architecture

**Data Gathering Procedures and Data Analysis Method**

The researchers will employ a systematic approach to collect a comprehensive dataset of facial images, ensuring a diverse range of conditions and characteristics to robustly test the enhanced facial recognition algorithm and compare it side by side with the Haar Cascade Algorithm with LBPH. The procedure is as follows:

1. Selection of Images:
    a. A total of 263 well-known individuals will be chosen to provide facial images. These participants will represent a diverse range of facial features and demographics, including various ages, genders, and ethnicities. Additionally, 12 non-face images will be included.
2. Environmental Variations:
    a. To ensure the dataset encompasses various real-world scenarios, images will be captured under different environmental conditions. This includes settings with complex backgrounds, dynamic lighting (such as varying light intensities and angles), and additional elements like glasses and varying hairstyles.
3. Image Collection
    a. Each selected famous personalities and non-face images will have a pair of different images taken from the internet. (Google, Facebook, Instagram, etc.) resulting in a total of 550 images. The dual-image approach per individual aims to capture slight variations in expression, angle, or lighting to test the algorithm's robustness and consistency

**Automated Testing Process**

- An automated testing script will be developed to handle the pairwise comparison of images. This script will ensure that each image is compared to every other image in the dataset.

- The script will iterate through the dataset, comparing each image (denoted as <image name>1 or <image name>2) to all other images except for the same file name. This will be repeated for every image in the dataset.
- A total of 301,950 comparison will be made, considering the 549 x 550 computation.

**Confusion Matrix Values**

- Face Detection
    - (TP): There's 1 face in the image.
    - (FP): There is more than 1 face in the image.
    - (TN): There's no face in the image.
    - (FN): There's 1 face in the image but it wasn't detected.
- Face Matching
    - (TP): Both faces are perfectly matched.
    - (FP): Different face but resulted to match or there's a face but not detected
    - (TN): Different face and returned as not match.
    - (FN): Same face but resulted as not match.

## 4) Result and Discussion:

After applying the face recognition library, RGB conversion, and logical process with filterng, the researchers conducted a testing and comparison between the Haar Cascade Algorithm and Enhanced Haar Cascade Algorithm

Figure 2 shows that after the enhancement with 128-dimensions, RGB conversion, and face filtering, the same image was tested and the result return a similarity percentage of 70.40% which is not match and there is no non-face detection, considering the acceptance threshold of 75%. This implies that the images are classified as "not matched" and is a True Negative since the images contain two different persons.

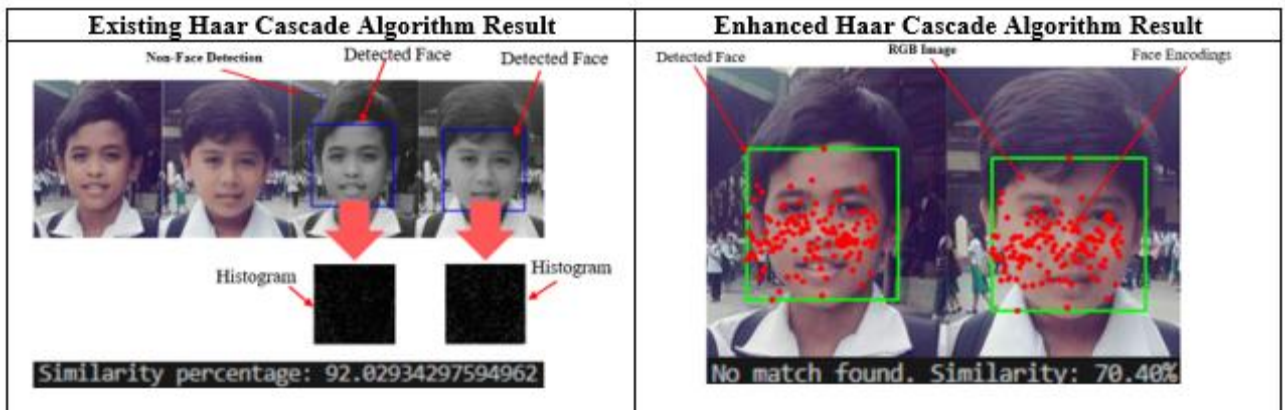

**Figure 2** Comparison between Haar Cascade and Enhanced Haar Cascade Algorithm

To differentiate the process and result of Haar Cascade and Enhanced Haar Cascade Algorithm visually the testing was performed using Eq. (1), RGB conversion, and face filtering from solution and showed in Table 1.

Table 2 shows the result of the 301,950 comparisons, it clearly indicates that the enhanced Haar Cascade algorithm significantly outperforms the standard version across all metrics except recall, where it maintains comparable performance. The stark improvement in accuracy and precision suggests that the enhancements have effectively reduced the number of false positives, leading to a more trustworthy face detection system. The high recall values in both algorithms imply that the detection coverage is comprehensive, ensuring that most faces are captured. The enhancement in the F1 Score for the enhanced algorithm underscores the balanced improvement across both precision and recall, making it a more robust and reliable option for practical applications.

**Table 2.** Confusion Matrix Result of Haar Cascade and Enhanced Haar Cascade Algorithm

|  | **Existing Haar Cascade Algorithm** | | | |
|---|---|---|---|---|
|  | **True Positive** | **False Positive** | **True Negative** | **False Negative** |
| Face Matching | 504 | 265234 | 12504 | 14 |
| Face Detection | 250500 | 33810 | 17424 | 216 |
|  | **Accuracy** | **Precision** | **Recall** | **F1 Score** |
| Face Matching | 4.674832 | 0.18966 | 97.2973 | 0.378583 |
| Face Detection | 88.73125 | 88.10805 | 99.91385 | 93.64031 |
|  | **Accuracy** | **Precision** | **Recall** | **F1 Score** |
| **TOTAL SCORE** | **46.70%** | **44.15%** | **98.61%** | **47.01%** |
|  | **Enhanced Haar Cascade Algorithm** | | | |
|  | **True Positive** | **False Positive** | **True Negative** | **False Negative** |
| Face Matching | 512 | 1186 | 282906 | 18 |
| Face Detection | 276150 | 8472 | 17328 | 0 |
|  | **Accuracy** | **Precision** | **Recall** | **F1 Score** |
| Face Matching | 99.57698 | 30.15312 | 96.60377 | 45.9605 |
| Face Detection | 97.19424 | 97.02342 | 100 | 98.48923 |
|  | **Accuracy** | **Precision** | **Recall** | **F1 Score** |
| **TOTAL SCORE** | **98.39%** | **63.59%** | **98.30%** | **72.23%** |
|  |  |  |  |  |
|  | **Accuracy** | **Precision** | **Recall** | **F1 Score** |
| Haar Cascade Algorithm | 46.70% | 44.15% | 98.61% | 47.01% |
| Enhanced Haar Cascade Algorithm | **98.39%** | **63.59%** | **98.30%** | **72.23%** |

The overall result of the 301,950 comparisons is categorized using the Accuracy, Precision, Recall, and F1 score summarized in Table 2.

### 5) Conclusion:

The enhancement achieved a 98.39% in accuracy, a 21.39% from the 46.70% to 77.00% baseline accuracy rate. Additionally, by adding face encoding with RGB images, the accuracy of facial matching increased substantially, as more detailed data was extracted, even when the images shared similar attributes. The enhanced algorithm has a 99.58% accuracy when comparing images with similar attributes. A 22.58% increase from the baseline accuracy rate of 46.70% to 77.00%. Furthermore, face detection accuracy improved after enhancing the baseline parameters with logical and filtering processes, reducing the detection of non-face objects in the images. It achieved a 97.19% accuracy rate, an increase of 8.46% from the previous face detection accuracy rate of 88.73%.